\documentclass[letterpaper, 10 pt, conference]{ieeeconf}  

\IEEEoverridecommandlockouts                              

\usepackage{graphicx}
\usepackage{amsmath} 
\usepackage{amssymb}  
\usepackage{multirow}
\usepackage{diagbox}
\usepackage{booktabs}
\usepackage{xcolor}
\usepackage{bm}

\definecolor{B}{RGB}{0, 128, 0}

\title{\LARGE \bf
WSCLoc: Weakly-Supervised Sparse-View Camera Relocalization \\
via Radiance Field
}

\author{Jialu Wang$^{1}$, Kaichen Zhou$^{1}$, Andrew Markham $^{1}$ and Niki Trigoni $^{1}$
\thanks{$^{1}$Jialu Wang, Kaichen Zhou, Niki Trigoni and Andrew Markham are with Department of Computer Science,
	University of Oxford,
	UK, 
    {\tt\small jialu.wang@cs.ox.ac.uk},
     {\tt\small rui.zhou@kellogg.ox.ac.uk}, 
    	{\tt\small niki.trigoni@cs.ox.\newline ac.uk}, 
    		{\tt\small andrew.markham@cs.ox.ac.uk}}
}

\begin{document}

\maketitle
\thispagestyle{empty}
\pagestyle{empty}

\begin{abstract}

Despite the advancements in deep learning for camera relocalization tasks, obtaining ground truth pose labels required for the training process remains a costly endeavor. While current weakly supervised methods excel in lightweight label generation, their performance notably declines in scenarios with sparse views. In response to this challenge, we introduce WSCLoc, a system capable of being customized to various deep learning-based relocalization models to enhance their performance under weakly-supervised and sparse view conditions. This is realized with two stages. In the initial stage, WSCLoc employs a multilayer perceptron-based structure called WFT-NeRF to co-optimize image reconstruction quality and initial pose information. 
To ensure a stable learning process, we incorporate temporal information as input. Furthermore, instead of optimizing SE(3), we opt for $\mathfrak{sim}(3)$ optimization to explicitly enforce a scale constraint. In the second stage, we co-optimize the pre-trained WFT-NeRF and WFT-Pose. This optimization is enhanced by Time-Encoding based Random View Synthesis and supervised by inter-frame geometric constraints that consider pose, depth, and RGB information. 
We validate our approaches on two publicly available datasets, one outdoor and one indoor.
Our experimental results demonstrate that our weakly-supervised relocalization solutions achieve superior pose estimation accuracy in sparse-view scenarios, comparable to state-of-the-art camera relocalization methods. We will make our code publicly available.



\end{abstract}

\section{Introduction}
\label{Introduction}
\begin{figure*}[htbp]
	\centering
	\includegraphics[width=0.95\textwidth]{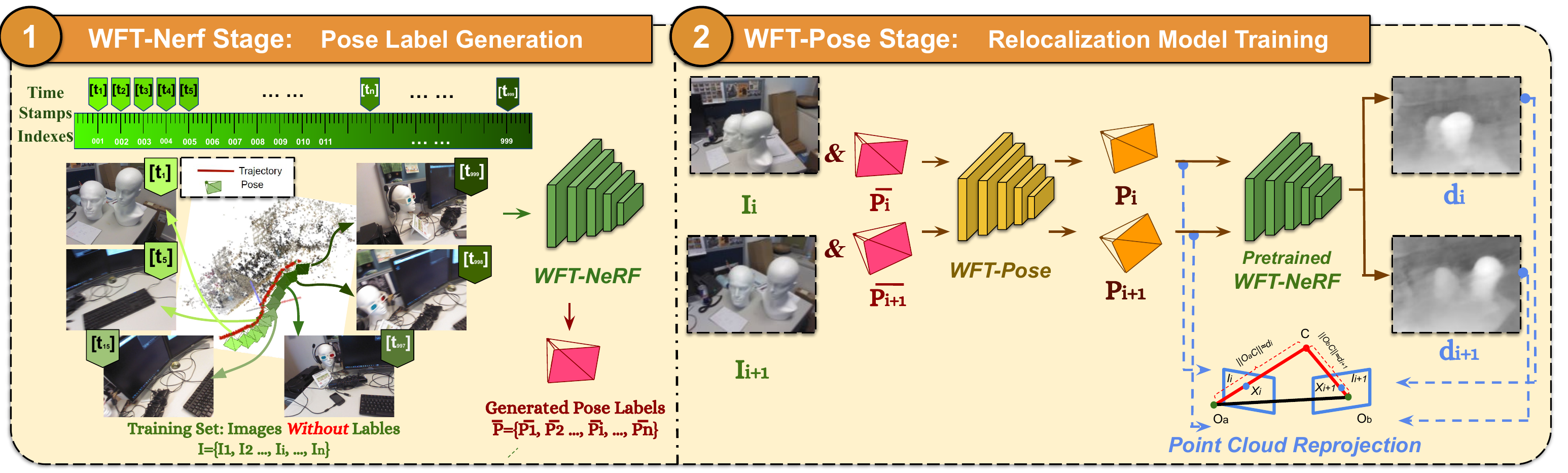}
	\caption{ \textbf{WSCLoc System Workflow.} In the WFT-NeRF Stage \textbf{(left)}, time encodings are generated for each image, and initial pose labels are obtained during the simultaneous training of the WFT-NeRF model. In the WFT-Pose Stage \textbf{(right)}, the training set is augmented using TE-based RVS (not shown in the figure). Consecutive frames are then fed into the target relocalization model in each iteration to calculate the pose loss and inter-frame geometric constraint loss. Finally, the relocalization model is trained by minimizing the overall loss.}
	\label{fig:ucloc0103}
\end{figure*}

Deep learning-based camera relocalization, which is essential in fields like autonomous driving and augmented reality (AR), continues to be a prominent area of research. This technology employs neural networks to implicitly learn a map of a given scene, allowing for the estimation of free-trajectories for a moving camera based on captured images. While state-of-the-art methods achieve high accuracy, they heavily rely on manually annotating dense-view images, posing two main challenges: Firstly, additional sensors like RGB-D cameras or LiDAR are often required. Secondly, this process can be time-consuming and involve significant manual effort, sometimes necessitating a dedicated team for up to a year \cite{maddern20171}.

To achieve weakly-supervised camera relocalization without heavy handcrafted labels, recent advancements leverage Structure-from-Motion (SfM) techniques \cite{kendall2015posenet, chen2021direct} to automatically generate labeled images from RGB data alone, eliminating the need for additional sensors. However, this approach still relies on dense-view images, which are computationally demanding and impractical for consumer-grade devices due to resource constraints. The primary challenges stem from the lack of depth information, leading to scale drift, and image distortions like motion blur, significantly impacting performance in sparse-view scenarios.

In this study, we introduce Weakly-Supervised Sparse-View Camera Relocalization via Radiance Field (WSCLoc), a system designed to achieve weakly-supervised camera relocalization without heavy handcrafted labels and achieve state-of-the-art relocalization performance in sparse view scenarios. Our approach comprises two stages. In the initial stage, we utilize neural radiance techniques to generate pose labels under sparse-view conditions through our WFT-NeRF model. In the following stage, we introduce WFT-Pose, leveraging the previously generated pose labels for relocalization network training. Additionally, performance is enhanced through inter-frame geometric constraints from the pre-trained WFT-NeRF model. During the inference stage, this pose estimator enables rapid pose estimation for unseen images within the current scene. Our contributions can be summarized as follows:

\begin{itemize}
	
\item We propose a WFT-NeRF model that employs neural radiance techniques to generate pose labels from highly sparse views, particularly in scenarios with free-trajectories and large-scale settings, without the need for additional sensors.

\item Leveraging the pose labels acquired in the initial stage, we propose WSCLoc, a system capable of being customized to various deep learning-based relocalization models to enhance their performance under weakly-supervised and sparse view conditions.

\item Comprehensive experiments are conducted on a publicly available large-scale outdoor dataset and an indoor dataset to validate the effectiveness of our model.

\end{itemize}

\section{Related Work}
\subsection{Weakly Supervised Camera Relocalization}
Camera relocalization determines a camera's precise 6-DoF pose based on input images. Recent deep-learning methods, like CNNs~\cite{wang2020atloc, kendall2015posenet, brahmbhatt2018geometry, chen2022dfnet, liu2023Nerf,zhou2021vmloc}, offer efficient and accurate pose estimation. However, these methods rely on precise pose labels, requiring additional sensors (e.g., RGB-D cameras, LiDAR, GPS) and time-consuming data preprocessing (e.g., depth map alignment)~\cite{zhou2022devnet}. In works such as \cite{kendall2015posenet, chen2021direct}, weakly supervised relocalization has been successfully achieved using the Structure-from-Motion (SfM) \cite{ullman1979interpretation} technique, but faces challenges in providing accurate pose labels due to issues like scale drift, image deformation in sparse-view scenarios.

\subsection{Sparse-View Camera Relocalization}
Recent advancements in camera pose estimation have yielded noteworthy contributions. For instance, RelPose \cite{zhang2022relpose} introduced a category-agnostic method for camera pose estimation, excelling particularly in rotation prediction but constrained by its ability to only predict rotations. Concurrently, SparsePose \cite{sinhasparsepose} pioneered camera pose regression followed by iterative refinement, while RelPose++ \cite{lin2023relpose++} decouples rotation estimation ambiguity from translation prediction through a novel coordinate system. However, these methods are confined to object-centric scenes. Moreover, \cite{dong2022visual} and PoseDiffusion \cite{wang2023posediffusion}, have shown promising results in camera pose estimation with very few annotated images. It's worth noting, though, that these methods still require additional sensors or dense-view structure-from-motion (SfM) reconstruction for image annotation. These research endeavors open new possibilities in camera pose estimation, yet underscore the need for further efforts to address data annotation requirements and applicability challenges in real-world scenarios.

\subsection{NeRF with Pose Estimation}
In recent research, there's a focus on eliminating the need for camera parameter preprocessing. Some methods like \cite{rosinol2022Nerf, sucar2021imap, zhu2022nice, yen2021iNerf, zhou2024dynpoint} rely on accurate camera poses from a SLAM tracking system or a pre-trained NeRF model. On the other hand, methods like \cite{wang2021Nerf, lin2021barf, jeong2021self, bian2023nope} go a step further by optimizing noisy camera poses during NeRF training. While these methods show promise for forward-facing datasets, they are often limited to handling forward-facing scenes or 360$^{\circ}$ object-centric unbounded scenes.

\begin{figure}[t]
	\centering
	   \includegraphics[width=0.98\linewidth]{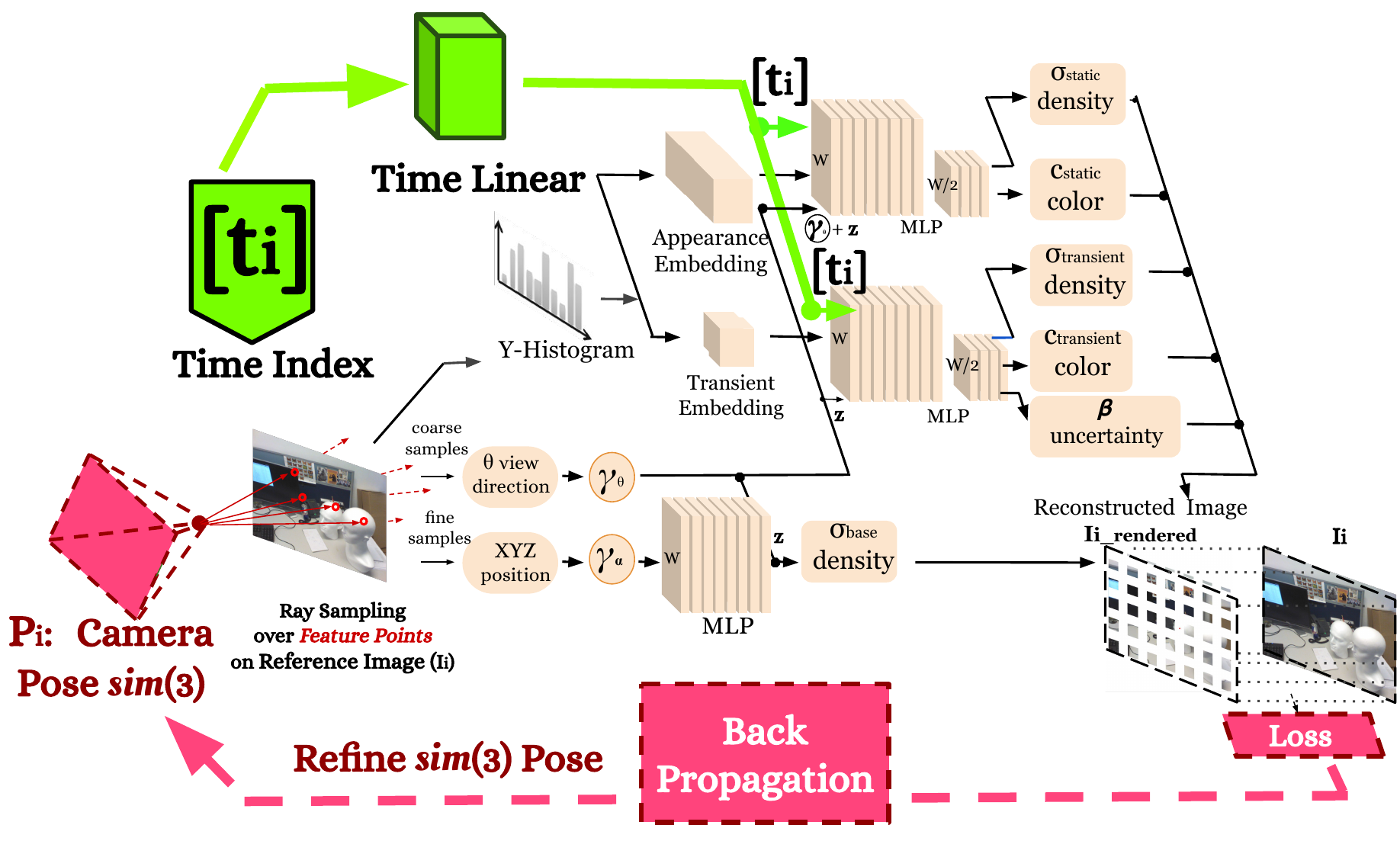}
	
	\caption{\textbf{Structure of WFT-NeRF.} During video capture, reference images are encoded with temporal information ($t_i$) using discrete time indices to minimize motion-related blurring. Grayscale levels in YUV are encoded for consistent exposure and appearance. Our NeRF training involves three sets of MLPs: 1. The base network estimates volume density and hidden state after coarse ray sampling. 2. Middle MLPs perform fine-ray sampling for appearance, estimating density and color. 3. Top MLPs handle fine-ray sampling for transient properties, estimating density, color, and uncertainty to filter transient objects. Losses between the rendered and reference images optimize pose during backpropagation, simultaneously optimizing NeRF and $\mathfrak{sim}(3)$ poses. Only the base network is used for testing.}
	\label{fig:SAT_Nerf_STRUCTURE02}
\end{figure}

\section {Methods}

The unsupervised camera relocalization is realized by WSCLoc with two steps. During the first step, WFT-NeRF is trained to generate initial pose information. During the second step, WFT-Pose is co-optimized with the WFT-NeRF to realize accurate pose estimation for unseen images.

\subsection{Weakly-supervised Free-Trajectory (WFT) NeRF } 
\label{Weakly-supervised Free-Trajectory (WFT) NeRF}

While methods like \cite{lin2021barf, wang2021Nerf} can refine noisy camera poses during NeRF training, they are inadequate for large-scale or free-trajectory scenes. The NeRF model from \cite{chen2022dfnet, wang2023f2} can handle such scenarios but lacks direct pose optimization capabilities. This is primarily because RGB images from monocular cameras can introduce scale drift, and artifacts may occur during NeRF model training due to distortions in images, such as rolling shutter effects, object deformations, or motion blur \cite{prisacariu2022direct}. To tackle these challenges, we introduced two key enhancements: 1. Explicit Scale Constraint to mitigate scale drifting. 2. Explicit Time Encoding to handle image distortions like boundary blurriness.

\subsubsection{Explicit Scale Constraint to Address Scale Drifting}
\label{sim3}
To mitigate scale drift, we introduce an additional scale factor $s$ using similarity transformations instead of Euclidean transformations ($SE(3)$) \cite{strasdat2010scale}. Given an initial $\text{SE}(3)$ pose, we represent the transformation as a $4\times4$ matrix (see Eq. (\ref{eq:5})), where $\textbf{R}$, $\textbf{t}$, and $s$ denote the rotation matrix, translation matrix, and scale factor, respectively. To maintain the estimated pose within the pose manifold during gradient-based optimization, we parameterize it using exponential coordinates.
\begin{equation}
	\mathfrak{sim}(3)= \begin{bmatrix}
		s\textbf{R} &\textbf{t} \\
		\textbf{0}^{T} &
		1\end{bmatrix} \in \mathbb{R}^{4*4}  ,
	\label{eq:5}
\end{equation}
To maintain the estimated pose within the pose manifold during gradient-based optimization, we parameterize it using exponential coordinates. Thus, we represent the transformation as a 7-dimensional $\text{sim}(3)$ vector, as shown in Eq. (\ref{eq:6}). Their exponential mapping relationships are described in Eq. (\ref{eq:7}, \ref{eq:8}), where the axis-angle representation is defined as \( \phi = \theta \mathbf{a} \),  \( \mathbf{a} \) is a unit direction vector, and \( \theta = \arccos \frac{\text{tr}(\mathbf{R})-1}{2} \) is the magnitude \cite{strasdat2010scale}.

\begin{equation}\small
    \mathfrak{sim}(3) = \left\{ \mathbf{P} \mid \mathbf{P} = \begin{bmatrix}
        \bm{\rho} \\ \bm{\phi} \\ \sigma
    \end{bmatrix} \in \mathbb{R}^{7} \right\} ,
    \label{eq:6}
\end{equation}

\begin{equation}
	s=e^{\bm{\rho}}, \mathbf{R} = \exp(\bm{\phi}^{\wedge}),  \mathbf{t} = J_s\bm{\rho},
	\label{eq:7}
\end{equation}

\begin{equation}
\begin{matrix}
\text{,where  } J_s = \frac{e^{\sigma} -1}{\sigma } + \frac{\sigma e^{\sigma}\sin {\bm{\theta}}+(1-e^{\sigma}\cos{\bm{\theta}})\bm{\theta}}{{\sigma}^2+
 {\bm{\theta}}^2}
 \\+(\frac{e^{\sigma}-1}{\sigma}-\frac{(e^\sigma\cos\bm{\theta}-1)\sigma+(e^\sigma\sin\bm{\theta})\bm{\theta}}{\sigma^2+\bm{\theta}^2})\mathbf{a}^{\wedge}\mathbf{a}^{\wedge},
 \end{matrix}
 \label{eq:8}
\end{equation}

In addition, considering that almost all NeRF methods utilize $R$ and $t$ to jointly render volume density but additionally employ $R$ to render color, we opted to independently optimize the translation and rotation components at different training rates, rather than equally optimizing the entire 7-dimensional $\mathfrak{sim}(3)$  vector.

\begin{equation}
	{\bar{\bm{\rho}} \bar{\bm{\phi}} \bar{\sigma}}=\underset{\bm{\bar\rho} \bm{\bar\phi} \bar\sigma \in \mathbb{R}^7}{\operatorname{argmin}} L(\bar{\mathbf{P}} \mid I, \bm{\Theta}),\label{eq:9}
\end{equation}

In the end, the problem of optimizing poses can be expressed using (Eq. (\ref{eq:8})). Let $\Theta$ represent the parameters of WFT-NeRF, $\bar{\mathbf{P}_i}= e^{\sigma_i}\exp (\bm{\phi_i}^{\wedge })\bar{\mathbf{P}_0} $ denote the estimated camera pose at the current optimization step $i$, $I$ represent the observed image, and $L$ be the loss used for training WFT-NeRF. Our goal is to determine the optimal pose starting from an initial estimate $\bar{\mathbf{P}_0}$ (in the sparse-view scenario, we obtain initial noisy pose estimates $\bar{\mathbf{P}_0}$ from SfM techniques). To achieve this, we employ gradient descent with the Adam optimizer to independently optimize the three components ($\bm{\rho}$, $\bm{\phi}$, and $\sigma$) of the $\mathfrak{sim}(3)$ pose using the photometric loss function $L$ of NeRF.

\subsubsection{Explicit Time Encoding to Addresses Boundary Blurriness}
\label{time encoding}
\begin{figure}[t]
	\centering
	   \includegraphics[width=0.8\linewidth]{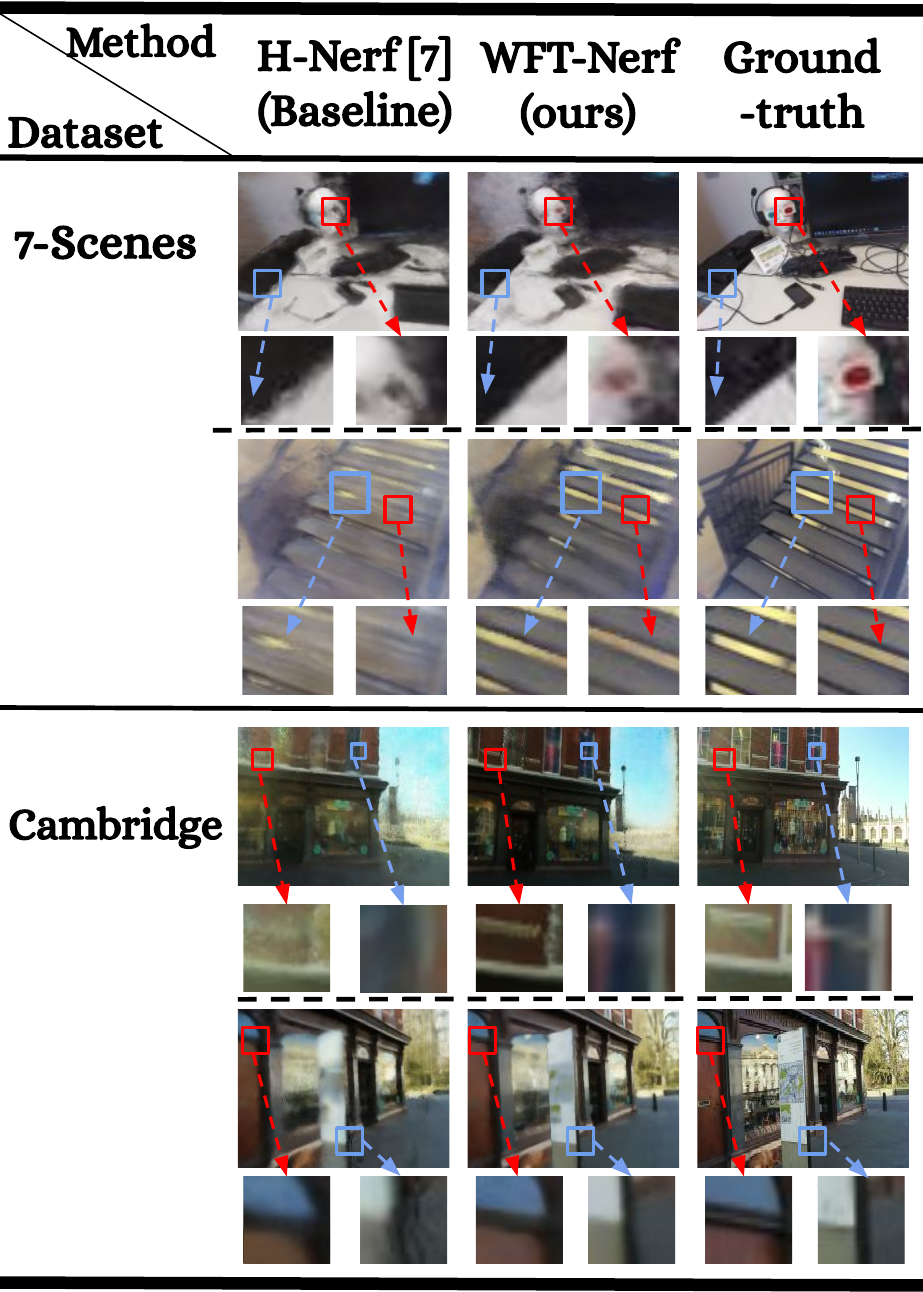}
	
	\caption{\textbf{Qualitative Comparison.} Large-scale or free-trajectories can introduce camera rolling shutter effects and motion blur, causing artifacts during NeRF model training and resulting in incorrect pose estimations. We mitigate this by explicitly encoding time indices for input images, rectifying deformed ground truth images, and enhancing robustness to motion blur with sharper object boundaries.}
	\label{fig:Time_encoding}
 \vspace{-0.6cm}
\end{figure}

Large-scale or free-trajectories scenarios can easily introduce camera rolling shutter effects and motion blur, which may lead to artifacts during the training of the NeRF model (Fig. \ref{fig:Time_encoding}), as demonstrated in \cite{prisacariu2022direct}, ultimately resulting in incorrect pose estimations.

To mitigate this issue, we explicitly encode discrete time indices for the input images (as shown in Fig. \ref{fig:SAT_Nerf_STRUCTURE02}). This introduces additional timestamp constraints to the pixel positions for each frame, effectively enforcing the camera's impact on objects to be smoother and reducing the boundary blurring caused by the camera's motion. As depicted in Fig. \ref{fig:Time_encoding}, this enables us to rectify deformed ground truth images, rendering sharper object boundaries, thereby enhancing robustness to motion blur.

\subsubsection{WFT-NeRF Structure Design}

The training process of WFT-NeRF is shown in Fig. \ref{fig:SAT_Nerf_STRUCTURE02}. To handle poses for large-scale or free-trajectories scenes, our WFT-NeRF structure is built upon a state-of-the-art NeRF model, showcased in \cite{chen2022dfnet}. In addition to incorporating our proposed techniques mentioned in \ref{time encoding} \ref{sim3}, we made two additional adjustments to further enhance its effectiveness: 1. To stabilize camera trajectory optimization in large scenes and ensure accurate registration, we replaced the original full positional encoding with a softer variant from \cite{chen2021direct}. This variant begins with a smooth signal and gradually shifts focus to learn a high-fidelity scene representation. We strongly recommend readers to refer to \cite{chen2022dfnet, chen2021direct} for more details. 2. To reduce computational costs, \cite{chen2022dfnet, yen2021iNerf, lin2021barf} employ partial ray sampling for faster training while maintaining accuracy. Our approach stands out by using SIFT feature extraction \cite{loweobject} to identify and match feature points between adjacent frames, enhancing rendering efficiency and performance in feature-rich areas for weakly supervised relocalization network training.

\subsection{Weakly-supervised Free-Trajectory Pose (WFT-Pose)}
In this section, we introduce WFT-Pose, a system capable of being customized to various deep learning-based relocalization models to enhance their performance under weakly-supervised and sparse view conditions. When applied to a target relocalization model, WFT-Pose optimizes it alongside a pre-trained WFT-NeRF model, integrating additional inter-frame geometric constraints and Time-Encoding Based Random View Synthesis (TE-based RVS) data augmentation to ensure accurate pose estimation for unseen images.

\subsubsection{Inter-Frame Geometric Constraints}
To compensate for the lack of scale information and mitigate scale drift issues in the input images from monocular cameras, we used depth maps generated by the WFT-NeRF to compute a point cloud loss based on KL divergence between adjacent frames. During this process, WFT-NeRF is pre-trained, and its weights are frozen.

During the training phase, for each pair of adjacent images (denoted as \{$I_i$, $I_{i+1}$\}, we perform feature matching to obtain a set of normalized coordinates of feature points \{$X_i$, $X_{i+1}$\} on the image plane by employing SIFT feature extraction \cite{loweobject} and nearest-neighbor search based on k-dimensional trees (k-d trees) \cite{friedman1975algorithm}.

Then we back-project the depth map \{$d_i$, $d_{i+1}$\} rendered by the pre-trained WFT-NeRF using the estimate poses \{$P_i$, $P_{i+1}$\} obtained by the target relocalization model, to point clouds \{$C_i$, $C_{i+1}$\} and optimise the relative pose between consecutive point clouds by minimising the point cloud loss $L_{pc}$ (see Eq. (\ref{eq:9}, \ref{eq:10})), where $P_{i+1,i} = P_{i+1}^{-1} \cdot P_i$ represents the relative pose that transforms point cloud $C_i$ to $C_{i+1}$.

\begin{align}
    L_{pc} & = \sum_{(i,\ i+1)}(C_{i+1},\ P_{(i+1,\ i)}\ C_i) \label{eq:9} \\
    & = \sum_{(i,\ i+1)}(d_{i+1}X_{i+1}P_{i +1},\ P_{(i+1,\ i)}\ d_iX_iP_i),
    \label{eq:10}
\end{align}

Since $d_i$ and $d_{i+1}$ inherently contain noise, and the feature points are quite sparse, we additionally introduce a KL divergence loss to further reduce the discrepancy between the distributions of the two frames. The final inter-frame geometric constraint loss (abbreviated as IF Loss) is shown in (Eq. (\ref{eq:11})), where $\mathcal{P}_{i+1}$ and $\mathcal{P}_i$ are the probability distributions of $C_{i+1}$ and $P_{(i+1,\ i)}\ C_i$, respectively. 

\begin{equation}
	L_{IF}  =  L_{pc}+ \sum_{(i,\ i+1)}(KL[\mathcal{P}_{i+1}, \mathcal{P}_i]) ,
	\label{eq:11}
\end{equation}

\subsubsection{Time-Encoding Based Random View Synthesis (TE-Based RVS)}
To enhance generalization, we adopt a strategy akin to \cite{moreau2022lens, chen2022dfnet} for generating additional training data by perturbing poses with NeRF. In contrast to previous methods, we utilize pretrained WFT-NeRF to introduce additional time encoding to these perturbed poses, thereby improving image sharpness and augmenting the effectiveness of data augmentation. TE-based RVS poses are generated around the training pose with a random translation and rotation noise of 0.2m and 10° respectively. For spaese-view scenarios with 20\% and 10\% of the train set images, we generate synthetic images at scales of 5, and 10 times the original number of images to ensure an equal scale of training images across all conditions.

\subsection{Weakly-supervised camera relocalization (WSCLoc)}

Expanding on our WFT-NeRF and WFT-Pose frameworks, we introduce WSCLoc, a versatile system designed to be capable of being customized to various deep learning-based relocalization models to enhance their performance under weakly-supervised and sparse view conditions. When integrated into a target relocalization model, WSCLoc first generates pose labels by training the WFT-NeRF model. Subsequently, it co-optimizes the target relocalization model with pre-trained WFT-NeRF to achieve accurate pose estimation for unseen images. The detailed workflow of WSCLoc is outlined below:

Training Phase of the WSCLoc (Fig. \ref{fig:ucloc0103}) is as follows: 

\begin{enumerate}
	\item \textbf{Image preprocessing} Given a series of RGB images, generate time encodings {$t_i, t_j, ...$} for each image, and perform feature matching between adjacent frames.
	
	\item \textbf{WFT-NeRF Stage: Pose Label Generation} Generating initial pose labels \{$ \bar{P_i}, \bar{P_{i+1}},...$\} during the simultaneous training of the WFT-NeRF model.
	
	\item \textbf{WFT-Pose Stage: Relocalization Model Training} First, augment the training set using TE-based RVS. Then, in each iteration, feed two consecutive frames $(I_i, I_{i+1})$ into the target relocalization model to calculate the pose loss and our inter-frame geometric constraint loss (Eq. (\ref{eq:11})). The relocalization model is trained by minimizing the overall loss.
	
\end{enumerate}

During the testing phase, we retain only the relocalization network for pose estimation.

\section{EXPERIMENTS}
\label{EXPERIMENTS}
\subsection{Relocalization Models and Evaluation Metrics}
\label{Relocalization models and Evaluation metrics}
To showcase the versatility of our system, we apply our WSCLoc to enhance two models for weakly supervised relocalization: DFNet \cite{chen2022dfnet}, a state-of-the-art NeRF-based model, and PoseNet \cite{kendall2015posenet}, the most traditional end-to-end relocalization model. In this study, we adhere to the same evaluation strategies and datasets as the original papers for all models to ensure a fair comparison. Specifically, we evaluate the median translation error (m) and median rotation error (degree) for all models.

\subsection{Datasets}
\label{Datasets}
We assess the effectiveness of our method using two widely-used datasets. The \textbf{7-Scenes Dataset} \cite{shotton2013scene} encompasses seven indoor scenes featuring RGB images, depth maps, and ground truth camera poses obtained from KinectFusion. The \textbf{Cambridge Landmarks Dataset} \cite{kendall2015posenet} comprises six large-scale outdoor scenes with RGB images, visual reconstructions of each scene, and 6-DoF ground truth camera poses reconstructed using SfM. Notably, due to inaccuracies in the 3D reconstruction of the STREET scene in Cambridge, consistent with prior studies \cite{brachmann2021visual, chen2022dfnet}, our experiments are conducted solely on the remaining five scenes.

\subsection{Implementation Details}
We trained all models under the dense-view condition (trained with 100\% of the train set images) and two sparse-view conditions (trained with 20\% and 10\% of the train set images), and then tested them on the complete (100\% images) test set. In the sparse-view experiments, we obtained the training set images by uniformly sampling every $5th$ and $10th$ image to achieve 20\% and 10\% of the training set, respectively, and then generated corresponding pose labels using SfM. It is worth noting that we chose the 20\% and 10\% image settings for experimentation because SfM fails to work properly in many scenes of these datasets when fewer than 10\% of the images are used. We utilized COLMAP \cite{schonbergerstructure}, one of the most popular Structure-from-Motion (SfM) tools, to generate train set pose labels for all scenes under sparse-view conditions. To ensure fairness in comparison, we maintained consistent hyperparameters across all models, following the settings used in the original papers. For training the WFT-NeRF models, we set the learning rate to 0.001. Training was conducted until convergence on a single Nvidia RTX-3090 GPU.

\subsection{WSCLoc Performance Evaluation}
\label{WSCLoc Performance Evaluation}
As WSCLoc is capable of being customized to various deep learning-based relocalization models to enhance their performance under weakly-supervised and sparse view conditions, we leverage our WSCLoc approach to enhance the DFNet \cite{chen2022dfnet} model, a state-of-the-art NeRF-based model, for weakly supervised relocalization, and compare its performance with that of the origional DFNet. Additionally, we conducted supplementary experiments on PoseNet \cite{kendall2015posenet}, a conventional end-to-end relocalization model, to demonstrate the versatility of our approach.

\subsubsection{Evaluation on DFNet \cite{chen2022dfnet} Model}
\label{DFNet Evaluation}
In this section, we leverage our WSCLoc approach to enhance the DFNet model for weakly supervised relocalization.In the WFT-NeRF training stage, in sparse-view scenarios, we simultaneously generate pose labels while training the WFT-NeRF. Subsequently, in the WFT-Pose stage, we integrate our NVS and inter-frame geometric constraints into the vanilla PoseNet model. We then benchmark against the origional DFNet \cite{chen2022dfnet} and conduct experiments on both the large-scale outdoor dataset (Cambridge Landmark \cite{kendall2015posenet}) and the indoor dataset (7scenes \cite{shotton2013scene}) to evaluate their performances.

\textbf{Evaluation on 7-Scenes Dataset} The experimental results on the 7-Scenes indoor dataset with complex trajectories are shown in Table \ref{tab1}. In sparse-view scenarios, as mentioned in Section \ref{Introduction}, the pose estimation accuracy from SfM tends to be noisier, resulting in lower relocalization accuracy for baseline methods (e.g., DFNet) compared to dense-view conditions. In contrast, our WSCLoc achieves relocalization accuracy comparable to dense-view conditions, even outperforming the benchmarks in some scenarios with only $10\%$ of the images (e.g., fire, redkitchen). In dense-view scenarios, our method achieves comparable results to the baseline. However, in certain cases, the baseline outperforms ours, possibly due to errors introduced by incorrect feature matching during FT Loss computation.

 \begin{table*}[h!]
		\centering
		\caption{\textbf{Performance Evaluation of WSCLoc on the DFNet Model \cite{chen2022dfnet} using the 7-Scenes DataSet \cite{shotton2013scene}}. We follow the original papers' evaluation protocols to ensure fairness. Specifically, we measure median translation and rotation errors in $m/^\circ$. }
		\label{tab1}
		\tabcolsep 10pt
		\begin{tabular}{@{}cc|ccccccc@{}}
			\toprule
			\%Imgs                 & Method      & Chess     & {Fire} & {Heads} & {Office} & {Pumpkin} & {Kitdhen} & {Stairs} \\ \midrule
			\multirow{2}{*}{100\%} & DFNet[7]     & 0.05/1.88 & 0.17/6.45     & \textbf{0.06}/\textbf{3.63}      & \textbf{0.08}/\textbf{2.48}       & 0.10/2.78        & 0.22/5.45        & \textbf{0.16}/3.29        \\
			& WSCLoc & 0.05/\textbf{1.83} & \textbf{0.13}/\textbf{4.80} & 0.07/3.88 & 0.13/2.99 & 0.10/\textbf{2.33} & \textbf{0.14}/\textbf{3.83} & 0.18/\textbf{3.20} \\\midrule
			\multirow{2}{*}{20\%} & DFNet[7]     & 0.18/6.30 & 0.41/10.12 & 0.21/14.04 & 0.82/9.43 & 0.40/8.51 & 0.51/12.82 & 0.79/12.20 \\
			& WSCLoc &      \textbf{0.05}/\textbf{1.84}     & \textbf{0.13}/\textbf{4.97}     & \textbf{0.07}/\textbf{3.90}      & \textbf{0.12}/\textbf{3.24}       & \textbf{0.10}/\textbf{2.46}        &      \textbf{0.14}/\textbf{3.95}            & \textbf{0.18}/\textbf{3.68}        \\\midrule
			\multirow{2}{*}{10\%} & DFNet[7]     & 0.19/9.23 &0.52/10.88  & 0.21/16.17 & 0.90/10.21 & 0.44/9.01 & 0.60/13.08 & 0.79/14.11 \\
			& WSCLoc & \textbf{0.05}/\textbf{1.86} & \textbf{0.17}/\textbf{5.33} & \textbf{0.13}/\textbf{7.19} & \textbf{0.14}/\textbf{3.78} & \textbf{0.11}/\textbf{2.90} & \textbf{0.14}/\textbf{3.76} & \textbf{0.22}/\textbf{4.71} \\ \bottomrule
		\end{tabular}
	\end{table*}

\textbf{Evaluation on Cambridge Dataset} We then evaluated our method on a more challenging large-scale outdoor dataset. The experimental results are summarized in Table \ref{tab2}. In the sparse-view scenario, all relocalization accuracies of the baseline noticeably degrade. In contrast, our WSCLoc method outperforms the baseline in all sparse-view scenarios, particularly achieving a 51\% and 33\% improvement over the averaged median translation and rotation errors in the 10\% images scenario compared to the baseline performance. In dense-view scenarios, our method achieves comparable results to the baseline, but in some cases, the baseline outperforms ours due to potential errors in feature matching during FT Loss computation.

\begin{table}[h!]
    \centering
    \caption{\textbf{{Performance evaluation of WSCLoc on the DFNet model \cite{chen2022dfnet} using the Cambridge DataSet \cite{kendall2015posenet}}.} We measure median translation and rotation errors in $m/^\circ$.}

    \label{tab2}
    \resizebox{\columnwidth}{!}{%
    \begin{tabular}{@{}cc|cccc@{}}
        \toprule
        \%Imgs                 & Method      &Kings& Hospital& Shop& Church \\ \midrule
        \multirow{2}{*}{100\%} & DFNet[7]     &0.73/2.37 & \textbf{2.00}/2.98 & \textbf{0.67}/\textbf{2.21} & \textbf{1.37}/4.03      \\
        & WSCLoc & \textbf{0.71}/\textbf{2.07}  & \textbf{2.00}/\textbf{2.79} & 1.02/2.92 &  1.52/\textbf{3.98} \\\midrule
        \multirow{2}{*}{20\%} & DFNet[7]     & 1.93/8.02 & 4.12/8.30 & 2.11/10.05 & 5.30/12.12\\
        & WSCLoc &       \textbf{1.31}/\textbf{2.59} & \textbf{2.94}/\textbf{3.61} & \textbf{1.38}/\textbf{5.85} & \textbf{3.16}/\textbf{6.73} \\\midrule
        \multirow{2}{*}{10\%} & DFNet[7]     & 2.39/8.77 & 4.30/9.78 & 3.47/12.20 & 7.02/14.77  \\
        & WSCLoc & \textbf{1.69}/\textbf{2.74} & \textbf{3.47}/\textbf{3.85} & \textbf{2.48}/\textbf{8.64} & \textbf{3.78}/\textbf{7.73} \\ \bottomrule
    \end{tabular}%
    }
\end{table}


\subsubsection{Evaluation on PoseNet \cite{kendall2015posenet} Model}
\label{PoseNet}
To demonstrate the versatility of our WSCLoc system, we applied it to the classic end-to-end relocalization model, PoseNet \cite{kendall2015posenet}, using the methodology outlined in Section \ref{DFNet Evaluation}. Following this, we conducted comparative experiments on the large-scale Cambridge dataset against the baseline PoseNet model. Table \ref{tab4} illustrates a notable decline in pose accuracy for the vanilla PoseNet with increasing view sparsity. In contrast, our WSCLoc method achieves results comparable to those obtained in dense-view scenarios. This indicates the effectiveness of WSCLoc in addressing the challenges posed by sparse views in camera relocalization tasks.

	\begin{table}[h!]
		\centering
		\caption{\textbf{Performance Evaluation of WSCLoc on the PoseNet Model \cite{kendall2015posenet} using the Cambridge DataSet \cite{kendall2015posenet}.} We measure median translation and rotation errors in $m/^\circ$.}
		\label{tab4}
		\begin{tabular}{cc|c}
			\toprule\% Imgs & Method & Avg. Pose Error \\
			\midrule \multirow{2}{*}{100 \%} & 
				PoseNet [2] & 2.04/6.23 \\
			&	WSCLoc (PN)
		 &\textbf{1.99}/\textbf{5.18} \\
			\midrule \multirow{2}{*}{20 \%} & PoseNet [2] &4.85/18.62\\
				&	WSCLoc (PN)
			& \textbf{2.47}/\textbf{6.33} \\
			\midrule \multirow{2}{*}{10 \%} &
			PoseNet [2] &8.02/20.38 \\
			&	WSCLoc (PN)
			&\textbf{2.82}/\textbf{7.41} \\
				\bottomrule
		\end{tabular}
	\end{table}

\subsection{WFT-NeRF Performance Evaluation}

\subsubsection{Quantitative Results for Pose Label Generation}
This section quantitatively demonstrates the performance of WFT-NeRF in generating pose labels in sparse-view scenarios. We obtained 20\% and 10\% of the train set images by uniformly sampling every $5th$ and $10th$ image, respectively, and then used SfM to generate pose labels in these two sparse-view scenarios. Subsequently, we compared the median translation and rotation errors between the pose labels generated by WFT-NeRF and the precise pose labels generated by SfM in the dense-view (100\% train set images) scenario. 

The experimental results are shown in Table \ref{tab3}. Compared to the precise poses obtained in the dense-view scenario, the pose error generated by SfM in both sparse-view scenarios with 20\% and 10\% of the images are significantly larger, and the errors increase as the sparsity increases. In contrast, the median translation and rotation errors of poses generated by WFT-NeRF are much closer to the ground truth poses botained in the dense-view scenario, demonstrating the ability of WFT-NeRF to effectively produce pose labels close to dense-view level even in highly sparse-view scenarios.

    \begin{table}[h!]
		\centering
		\caption{\textbf{The error between generated pose labels under sparse-views and those generated under dense-view (100\%) by SfM.} We measure median translation and rotation errors in $m/^\circ$.}
		\label{tab3}
		\begin{tabular}{cc|cc}
			\toprule Data Set & \% Imgs & SfM & WFT-NeRF \\
			\midrule \multirow{2}{*}{ 7 Scenes} & 20 \% & 1.52/7.11 & \textbf{0.05}/\textbf{0.31}\\
			& 10 \% & 1.64/7.83 & \textbf{0.07}/\textbf{0.41}\\
			\midrule \multirow{2}{*}{ Cambridge } & 20 \% &1.16/9.12& \textbf{0.02}/\textbf{0.53} \\
			& 10 \% &1.70/13.36 & \textbf{0.06}/\textbf{0.72} \\
			\bottomrule
		\end{tabular}
  \vspace{-0.6cm}
	\end{table}
 
\subsubsection{QualitativeResults for Image Rendering}
This section qualitatively demonstrates the performance of WFT-NeRF in rendering images in large-scale and complex trajectories scenes. We employed DFNet's Histogram-assisted NeRF \cite{chen2022dfnet}, a state-of-the-art NeRF variant capable of operating in large-scale or free-trajectory scenes, as our baseline. We conducted experiments on both the large-scale Cambridge dataset and the 7-Scenes dataset with complex trajectories. Figure \ref{fig:Time_encoding} qualitatively demonstrates the rendering performance of our WFT-NeRF and the baseline in extremely sparse-view scenarios on these challenging datasets. The baseline exhibits blurry boundaries in rendered images when significant changes occur in camera poses. In contrast, our WFT-NeRF produces clearer boundaries, which is beneficial for subsequent tasks such as WSCLoc implementation of TE-based random view synthesis and depth-map rendering. The PSNR results of NeRF can be found in the ablation study.

\begin{figure}[t]
	\centering
	\rule{\linewidth}{0pt}
	\includegraphics[width=0.85\linewidth]{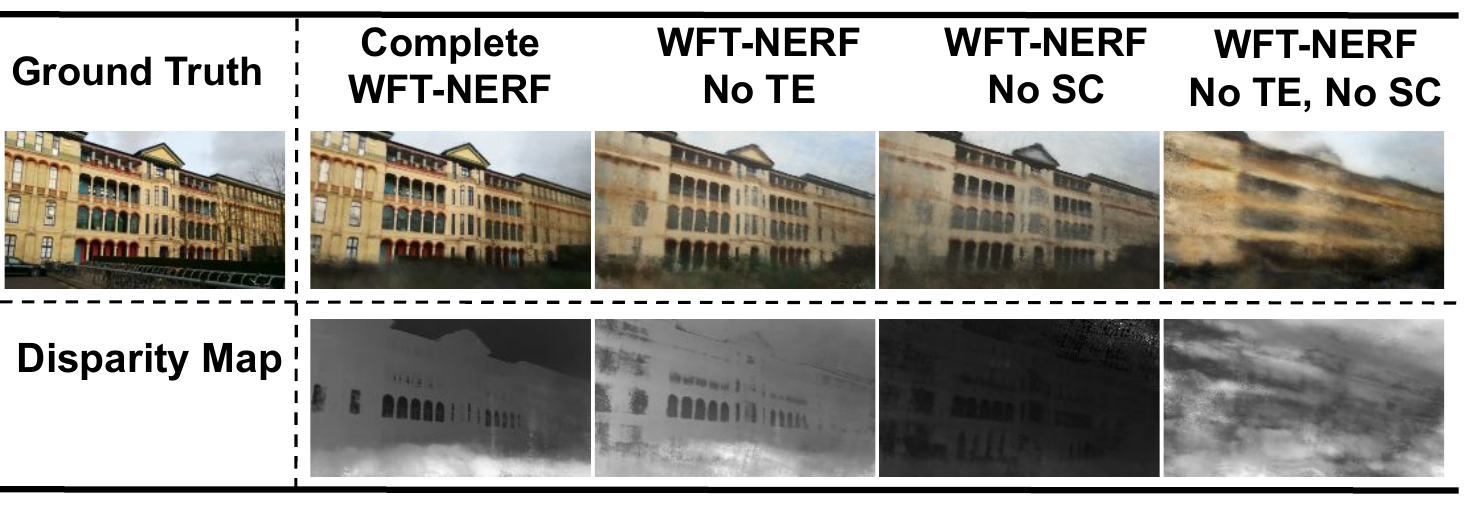}
	
	\caption{\textbf{Qualitative Evaluation of WFT-NeRF Performance} (Hospital scene in the large-scale Cambridge dataset under 10\% sparse-view conditions). Here, we demonstrate the impact of our full WFT-NeRF model benefiting from explicit scale constraint (SC) and Time Encoding (TE). Removing TE alone results in less clear image boundaries. Removing SC only leads to blurry images due to noisy pose labels generated by SfM in sparse-view scenarios. Removing both TE and SC severely degrades the quality of rendered images.}
	\label{fig:PSNR}
 \vspace{-0.5cm}
\end{figure}

\subsection{Ablation Study}
\label{Ablation Study}
\subsubsection{Effectiveness of Explicit Scale Constraint and Time Encoding of WFT-NeRF}
\label{Ablation Study-1}
We evaluated the effectiveness of Explicit Scale Constraint (SC) and Time Encoding (TE) in WFT-NeRF under highly sparse-view conditions on the Cambridge Dataset. We trained the model using 10\% of the train set images and tested the rendering performance of WFT-NeRF on the complete test set. As shown in Fig. \ref{fig:PSNR}, the results of qualitative experiments demonstrate that the Explicit Time Encoding effectively improves the clarity of object boundaries in rendering. From the quantitative results in Table \ref{tab5}, we found that in sparse-view scenarios, the Explicit Scale Constraint provides greater assistance in improving PSNR by generating more accurate poses.

\subsubsection{Effectiveness of TE-based RVS and inter-frame geometric constraints of WFT-Pose}
We also evaluated the effectiveness of TE-based RVS and inter-frame geometric constraints (IF Loss) in WFT-Pose under the same experimental settings as \ref{Ablation Study-1}. As shown in Table \ref{tab5}, we found that in highly sparse-view scenarios, the inter-frame geometric constraints provides less assistance in reducing pose errors compared to TE-based RVS. We attribute this to the sparsity of feature points matched between adjacent frames in highly sparse-view scenarios, which weakens the geometric constraints for relocalization. In future work, we will consider using deep learning-based feature matching methods to further optimize its performance.

    \begin{table}[h!]
		\centering
		\caption{\textbf{Ablation Study 10\% Imgs on Cambridge Dataset}. We measure median translation and rotation errors in $m/^\circ$.}
		\label{tab5}
		\begin{tabular}{@{}ccc||ccc@{}}
			\toprule
			\multicolumn{3}{c||}{WFT-NeRF}                                                & \multicolumn{3}{c}{WFT-Pose}                                                       \\ \midrule
			\multicolumn{1}{c|}{SC}       & \multicolumn{1}{c|}{TE}           & PSNR & \multicolumn{1}{c|}{TE-based RVS}          & \multicolumn{1}{c|}{IF Loss}      & Avg. Pose Error \\ \midrule
			\multicolumn{1}{c|}{}             & \multicolumn{1}{c|}{}             & 16.20 & \multicolumn{1}{c|}{}             & \multicolumn{1}{c|}{}             & 3.44/6.20    \\
			\multicolumn{1}{c|}{$\checkmark$} & \multicolumn{1}{c|}{}             & 19.87 & \multicolumn{1}{c|}{$\checkmark$} & \multicolumn{1}{c|}{}             & 2.88/5.90    \\
			\multicolumn{1}{c|}{}             & \multicolumn{1}{c|}{$\checkmark$} & 16.80 & \multicolumn{1}{c|}{}             & \multicolumn{1}{c|}{$\checkmark$} & 2.92/6.03    \\
			\multicolumn{1}{c|}{$\checkmark$} & \multicolumn{1}{c|}{$\checkmark$} & \textbf{20.95} & \multicolumn{1}{c|}{$\checkmark$} & \multicolumn{1}{c|}{$\checkmark$} & \textbf{2.85}/\textbf{5.74} \\ \bottomrule
		\end{tabular}
  \vspace{-0.7cm}
	\end{table}

\section{CONCLUSIONS}

In summary, traditional deep learning-based camera relocalization models heavily rely on manually annotating dense-view images. While existing weakly-supervised methods excel in lightweight label generation, their performance significantly degrades in sparse-view scenarios. To address these challenges, we introduce WSCLoc, a system capable of being customized to various deep learning-based relocalization models to enhance their performance under weakly-supervised and sparse view conditions. WSCLoc first generates pose labels by training our WFT-NeRF, then co-optimizes the target relocalization model with the pre-trained WFT-NeRF to achieve accurate pose estimation for unseen images. Our innovations include explicit scale constraint and time encoding of our WFT-NeRF, as well as Time-Encoding Based Random View Synthesis and Inter-Frame Geometric Constraints of our WFT-Pose. Experimental results on diverse datasets demonstrate that our weakly-supervised solutions achieve state-of-the-art accuracy performance in sparse-view scenarios.

\end{document}